\documentclass[10pt,twocolumn,letterpaper]{article}

\usepackage{iccv}              

\usepackage{multirow} 
\usepackage{pifont}
\usepackage{colortbl}
\usepackage{float}
\usepackage{xcolor}
\usepackage{booktabs}
\usepackage[accsupp]{axessibility}  
\definecolor{iccvblue}{rgb}{0.21,0.49,0.74}
\usepackage[pagebackref,breaklinks,colorlinks,allcolors=iccvblue]{hyperref}

\def\paperID{4016} 
\def\confName{ICCV}
\def\confYear{2025}

\title{

TextSSR: Diffusion-based Data Synthesis for Scene Text Recognition

}

\author{
Xingsong Ye$^{1}$, Yongkun Du$^{1}$, Yunbo Tao$^{1}$, Zhineng Chen$^{2}$\thanks{Corresponding author.}\\
$^1$College of Computer Science and Artificial Intelligence, Fudan University, China\\
$^2$Institute of Trustworthy Embodied AI, Fudan University, China\\
  {\tt\small \{xsye20, zhinchen\}@fudan.edu.cn, \{ykdu23, ybtao24\}@m.fudan.edu.cn }
}

\begin{document}
\maketitle
\begin{abstract}

Scene text recognition (STR) suffers from challenges of either less realistic synthetic training data or the difficulty of collecting sufficient high-quality real-world data, limiting the effectiveness of trained models. Meanwhile, despite producing holistically appealing text images, diffusion-based visual text generation methods struggle to synthesize accurate and realistic instance-level text at scale. To tackle this, we introduce \textbf{TextSSR}: a novel pipeline for \textbf{S}ynthesizing \textbf{S}cene \textbf{Text} \textbf{R}ecognition training data. TextSSR targets three key synthesizing characteristics: accuracy, realism, and scalability. 
It achieves accuracy through a proposed region-centric text generation with position-glyph enhancement, ensuring proper character placement. It maintains realism by guiding style and appearance generation using contextual hints from surrounding text or background. This character-aware diffusion architecture enjoys precise character-level control and semantic coherence preservation, without relying on natural language prompts. Therefore, TextSSR supports large-scale generation through combinatorial text permutations. Based on these, we present TextSSR-F, a dataset of 3.55 million quality-screened text instances. Extensive experiments show that STR models trained on TextSSR-F outperform those trained on existing synthetic datasets by clear margins on common benchmarks, and further improvements are observed when mixed with real-world training data. Code is available at \url{https://github.com/YesianRohn/TextSSR}.

%

\end{abstract}    
\section{Introduction}

\begin{figure}
    \centering
    \includegraphics[width=1\linewidth]{./imgs/overview.pdf}
    \caption{\textbf{Top}: Text instances detected by an end-to-end STR model~\cite{li2022pp}, and synthesized by our TextSSR based on the detected examples. \textbf{Bottom}: Accuracy of MAERec~\cite{jiang2023revisiting} across common benchmarks. \textcolor[HTML]{FAE4D5}{Peach} indicates training on ST~\cite{st} and TextOCR~\cite{singh2021textocr}, while \textcolor[HTML]{C8DCED}{blue} denotes that TextSSR-F is additionally included as training data.}
    \label{fig:overview}
\end{figure}

Scene text images are a unique type of image captured in the wild, they differ from general images in the presence of text, which appears with diverse sizes, fonts, and often with varying degrees of distortion and occlusion. However, these texts usually contain high-level semantic information closely related to the captured scene, making STR a crucial task for complex scene understanding, image search, etc.

Recent STR practices~\cite{jiang2023revisiting,du2023cppd,du2024igtr} have shown that besides advanced modeling techniques, high-quality training data also plays a critical role. Training STR models also follows the scaling law from the data perspective \cite{rang2024empirical} and current data size is far from saturated even for English. Enlarging the size of real-world datasets is a feasible way to further enhance recognition performance. However, it is challenging to collect real-world scene text images on an even larger scale. On one hand, diversified data collection is costly and time-consuming. On the other hand, real-world scenes predominantly contain high-frequency, semantically meaningful words, while low-frequency words are difficult to collect. Synthesizing high-quality text images seems to be an effective alternative. However, research on this direction is somewhat laggard. The two most popular synthetic datasets, MJ~\cite{mj} and ST~\cite{st}, are constructed based on traditional rendering techniques nearly a decade ago. Although they can be easily scaled up in quantity, recent studies~\cite{du2023cppd,du2024svtrv2} indicate that there are significant accuracy gaps between models trained on MJ\&SJ and Union14M-L~\cite{jiang2023revisiting}, mainly because MJ\&SJ fail to comprehensively describing the challenge of real-world text instances.

Recently, diffusion models have shown impressive performance in generating text-rich images~\cite{glyphcontrol, anytext, textdiffuser, textdiffuser2, udifftext}. However, most of them focus on generating aesthetic text images holistically, e.g., posters. Using data generated by these methods for STR model training encounters several issues. First, these methods are designed for the full image level generation, when switching to synthesizing text in instance-level regions, it would be less effective and ensuring the generation accuracy becomes challenging. Second, current generation methods mainly rely on well-refined natural language as prompts. They produce overly homogeneous and unrealistic text instances with some probability, which is not suitable as training data. Moreover, constructing a diverse set of prompts is inherently difficult, which hinders the ability to generate data at scale. 

Nevertheless, given the exceptional performance of diffusion models across various tasks, it is anticipated that there are diffusion-based techniques for producing satisfactory training data for STR. Before proceeding, we first outline three key characteristics that an improved data synthesis method should exhibit:

\begin{itemize}
\item \textbf{Accurate}: The synthesized text content should match the given text, ensuring that the synthesized words are readable and free from character misplacement or duplication errors. This is fundamental for maintaining the correctness of the trained STR models.
\item \textbf{Realistic}: The generated text instances should visually resemble real-world text, simulating scene conditions in the wild. This includes being harmoniously blended into the background, ensuring consistent generation at different sizes and capturing the diversity of text presentation, especially in complex scenes. Achieving realism is essential for enhancing the performance of text recognizers.
\item \textbf{Scalable}: The method should support extensive training data production. In other words, abandon complex control inputs and design processes, and enable large-scale data generation by leveraging easily accessible and processable resources only.
\end{itemize}

\begin{figure}
    \centering
\includegraphics[width=0.9\linewidth]{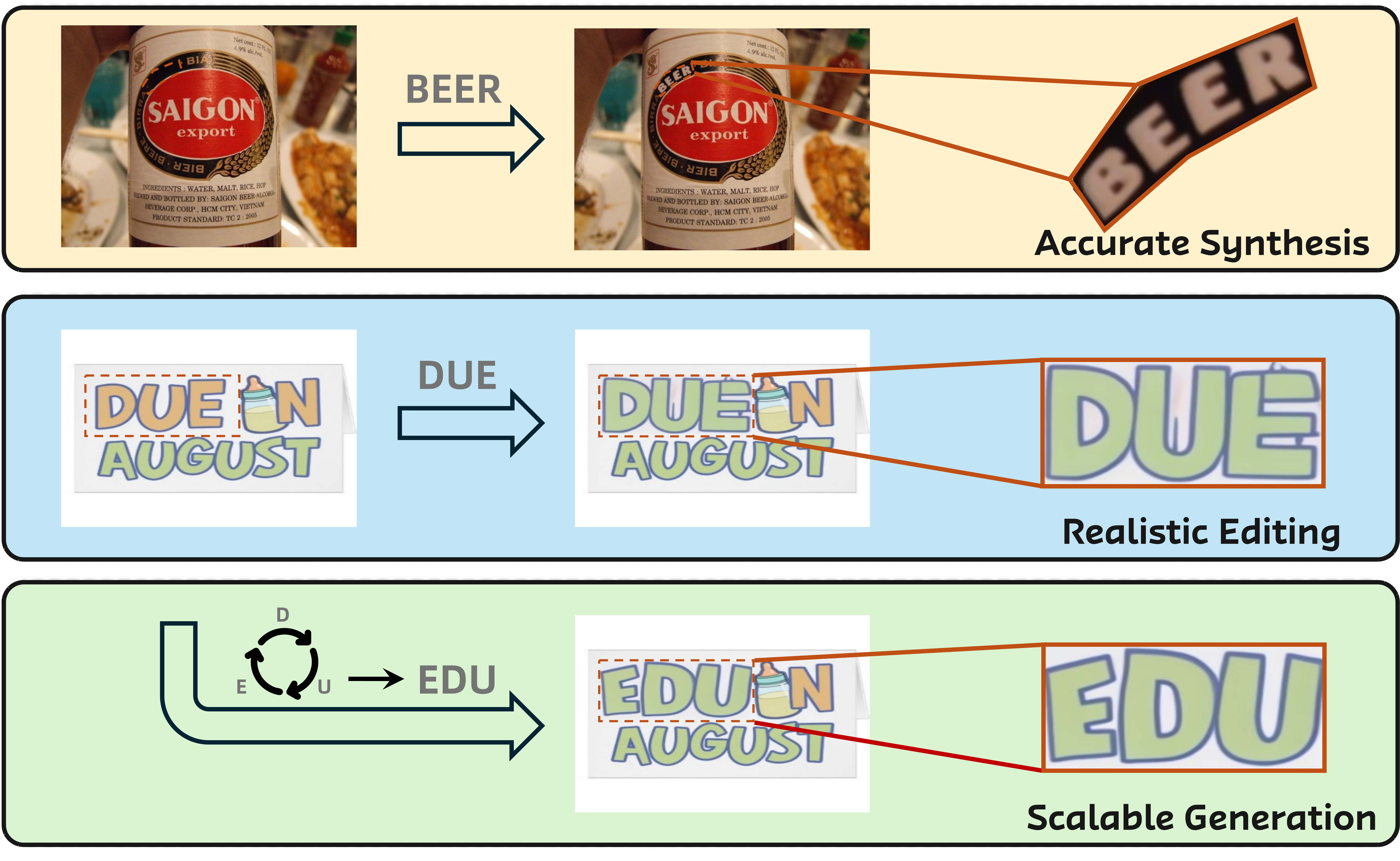}
    \caption{Showcases of TextSSR's capacity to synthesize accurate, realistic and scalable text instances.}
    \label{fig:showcases}
\end{figure}

To satisfy these characteristics, we propose \textbf{TextSSR}, a novel diffusion-driven pipeline focusing on \textbf{S}ynthesizing training data for \textbf{S}cene \textbf{Text} \textbf{R}ecognition. Our approach introduces the following innovations to overcome persistent limitations in text synthesis. First, we implement a region-centric text generation. It employs end-to-end OCR model to acquire text location and size. Moreover, we devise a position-glyph enhancement that further resolves character duplication and positional disorder, ensuring precise text placement and arrangement. Second, we leverage the detected text or only background near the designated region as contextual hints for font style and appearance generation, guiding the generated text to match real-world scenarios better. Third, we design a character-aware diffusion architecture that employs glyph priors rather than natural language prompts as injected conditions. It enables precise character-level control while maintaining full-word semantic coherence. To expand the generation quantity, we use detected text to generate a series of anagrams by rearranging characters within the original bounding boxes, thus supporting large-scale generation. As a result, TextSSR achieves accurate, realistic, and scalable text instance synthesize, as illustrated in Fig.~\ref{fig:showcases}.

Through these attempts, we construct the TextSSR-F dataset with 3.55 million quality-screened samples, which is automatically validated by checking their recognitions and pseudo labels. Extensive experiments demonstrate the superiority of TextSSR-F over existing synthesis data, as well as its complementarity with real-world training data.

Our contributions are summarized as follows:

\begin{itemize}
    \item We propose TextSSR, a diffusion-based text synthesis method centered on seamlessly blending text instances into real scenes. It develops a region-centric processing that enables accurate instance-level generation.
    \item TextSSR fully exploits positional and glyph priors of text, achieving accurate, diverse and realistic text synthesis by using only the detected text as input. Moreover, it can be easily scaled up and we construct the TextSSR-F dataset with 3.55 million text instances. 
    \item Extensive evaluations validate that STR models trained on TextSSR-F surpass their counterparts trained on existing synthetic data by 3.1\% on common benchmarks, and even better STR models can be obtained when further combined with real-world training data, highlighting the utility of the proposed TextSSR.  
    
\end{itemize}

\section{Related Work}

\subsection{Scene Text Recognition}

STR researches can be categorized from both the model and the data perspectives. From a model point of view, early approaches employ the CTC-based pipeline~\cite{crnn, zhai2016chinese}. They used CNN for visual feature extraction, followed by RNN for sequence modeling, then with CTC loss applied as a constraint. Subsequently, attention mechanisms gradually gained attention in the STR field~\cite{sheng2019nrtr, luo2019moran, shi2018aster, li2019show}. Inspired by developments in the NLP field, many methods~\cite{fang2021abinet, bautista2022scene, fang2022abinet++, wang2022petr, wei2024image, zheng2024cdistnet,Xu_2024_CVPR_OTE} further refined recognizers from a language modeling perspective. While some approaches~\cite{du2022svtr, zhang2024self, du2024svtrv2} continued to explore solutions based on a single vision model and CTC decoding, due to their fast inference nature. Accuracy was steadily improved following these efforts.

From a data perspective, the limited availability of well-annotated real-world training data led to a practical STR training and evaluation protocol: training solely on synthetic data, or a combination of synthetic and real data, followed by evaluation on real-world benchmarks. Besides the six common benchmarks~\cite{{iiit, svt, ic13, ic15, svtp, cute}}. Union14M-L~\cite{jiang2023revisiting} aggregated publicly available real-world datasets (including TextOCR~\cite{singh2021textocr}) and created a real-world training dataset with nearly 4 million instances. However, recent studies on the OCR scaling law~\cite{rang2024empirical} suggested that even this substantial volume had not reached the saturation point. This observation implies that STR models can still benefit from additional training data, and high-quality synthetic data could be a feasible supplement, which motivates our study.

\subsection{Scene Text Synthesis}

Early synthesis methods~\cite{mj, st, unrealtext} employed rule-based methods to overlay text onto background with transformations simulating real-world appearance. To be more realistic, later research shifted toward deep learning-based generation methods. Among these, GAN-based scene text editing~\cite{wu2019editing, yang2020swaptext, qu2023exploring} methods aimed to expand data by editing text directly within specified regions, but they were limited by the lack of real paired training data. Recent advances had leveraged diffusion models for both editing~\cite{ji2023improving, santoso2024manipulating, zeng2024textctrl} and synthesis, with fine-grained control over glyphs~\cite{DiffUTE, glyphcontrol}, as well as positioning at word~\cite{anytext} and character~\cite{textdiffuser, udifftext, DreamText} levels. The latest methods~\cite{textdiffuser2, SceneVTG} incorporate large language or multi-modal models to enhance performance. However, these approaches primarily targeted visually appealing generation for aesthetic purposes, rather than generating data for better STR models. While some work has explored diffusion-based text image synthesis. For example, CTIG-DM~\cite{zhu2023conditional} employed character sequence as the condition for generation. However, it is less controllable and faced with the dilemma of diversity-realism tradeoff, limiting quality of the generated text. SceneVTG~\cite{SceneVTG} developed a first-erasing-then-synthesizing pipeline. It requires a pre-erased model to remove existing text completely, which is difficult to guarantee in large-scale synthesizing scenarios.  

\section{Method}

To achieve accurate, realistic, and scalable text instance generation, TextSSR proposes a four-step pipeline (detailed in Fig.~\ref{fig:model}(b)). To handle text instances with varying sizes and locations, we first focus on text region processing, which will be discussed in Sec.~\ref{sec:3-1}. To enhance accuracy through fine-grained extraction and decoupling of text glyphs and positions, Sec.~\ref{sec:3-2} and Sec.~\ref{sec:3-3} will detail the specific module design and training strategy, respectively. Sec.~\ref{sec:3-4} introduces our data extension method, achieving scalability in data synthesis at scale.

\begin{figure*}
    \centering
    \includegraphics[width=0.95\linewidth]{imgs/model_pipeline.pdf}
    \caption{Overall architecture of TextSSR's (a) generative model and (b) synthesis pipeline. (a) Given a scene text instance (marked in \textcolor[HTML]{FF0000}{red}) and its content (e.g., ``kills"), it is preprocessed into the smallest local image containing the text. A text-size mask is then applied to identify non-text regions, while the text content is rendered as priors at both word and character levels. These different conditions are subsequently injected into a diffusion model as training guidance.
    (b) An end-to-end OCR model extracts the text location and content within the image (Step 1). All possible text words are generated through anagram (Step 2). The generative model then synthesizes text instances in batches (Step 3), and finally, a SOTA STR model filters the samples, retaining only the correctly recognized instances for use (Step 4).}
    \label{fig:model}
\end{figure*}

\subsection{Text Region-centric Processing}
\label{sec:3-1}

Instance-level text generation presents significant challenges, as we must handle varying sizes, diverse potential positions, and determine how to select appropriate surrounding information for effective prompting. Synthesis methods that edit the entire image are not suitable for this task. Such approaches create an imbalance, and small text regions lack sufficient pixels for accurate reconstruction. Furthermore, information from the entire scene is often redundant for text region synthesis. For example, in Fig.~\ref{fig:model}(a), the text ``kills" to be synthesized shares a similar style and context with the nearby upper text and background, while the region farther away is inconsistent in appearance.

Therefore, we propose a generative scheme based on a locality assumption—\textit{text styles are more likely to correlate with adjacent text and background information}. This method focuses generation at the region level, synthesizing the text using only surrounding local context rather than the entire scene. Specifically, we use the annotated text box to extract the smallest enclosing square around the text region and resize it to a fixed size (e.g., 256). For practical considerations, we uniformly employ an end-to-end OCR tool~\cite{li2022pp} to obtain text instance size, position, and textual labels. It well reflects real-world scenarios where annotations are often unavailable.

\subsection{Fine-grained Condition Mechanisms}
\label{sec:3-2}

Visual text rendering needs to consider two key fine-grained features: glyph structure and positional arrangement, both at word (holistic) and character (individual) levels. Word-level features concern entire word appearance and region placement, while the character-level address individual characters and their relative ordering. To guide the generative process, we implement the following approach:

\noindent\textbf{Word-level Position and Glyph Guidance.} 
For a text $t$ with position $P_t$ in the original image, we extract a square region $I_s$ where $P_t$ becomes $P_t'$ (Fig.~\ref{fig:model}(a), red dashed rectangle). This allows us to establish our desired word-level position guidance. Then we create a binary mask $I_M$ that separates text (0) from non-text (1) regions. The size of $I_M$  is identical to $I_s$, indicating where the text is present.  For glyph guidance, we render the text in a standard font, applying an affine transformation to align it with the position and orientation of $P_t'$ (resulting in $I_G$). As noted in previous work~\cite{DiffUTE, anytext, glyphcontrol}, \textit{when we close the door on the original styled text, we must at least provide a window with a basic visual style of the text at that location}.

\noindent\textbf{Character-level Position and Glyph Guidance.}
Only word-level information limits the model's understanding to the words, as we observe issues such as repeated or incorrect characters when using only word-level features. 
We address this by encoding character-level ordering through pixel values, enabling the model to learn spatial relationships between characters.
Specifically, we define a maximum character length $L$. For characters from position 1 to $l$ ($l \leq L$, with truncation if $l > L$ and zero-padding in channels $l+1$ to $L$ if $l < L$), the $i$-th character is rendered into the $i$-th channel of an $L$-channel image $I_g$.  Each character glyph is rendered with a pixel intensity of $i \times \left\lfloor \frac{P_M}{L} \right\rfloor$, where $P_M$ denotes the maximum number of pixels allocated to distinguish a character from the background. We represent the glyph image tensor as $I_g \in \mathbb{R}^{L \times S \times S}$, where $L$ is the number of characters and $S$ is the image size for each character. This creates position-prior pixel values that serve as positional encodings to help the model distinguish character information at each specific position.

Vision Transformer (ViT)~\cite{dosovitskiy2020image} is employed as the Glyph Encoder $G_{\phi}$ to accommodate input $I_g$  and output features compatible with diffusion model's condition inputs:
\begin{equation}
F_g = G_\phi (I_g) \in \mathbb{R}^{(N + 1) \times D}
\label{glyph}
\end{equation}
Here, $N$ represents the number of patches, the additional one corresponds to a class token, and dimension $D$ is set to match the input requirements of the diffusion process.

\subsection{Adapting to Stable Diffusion}
\label{sec:3-3}

We leverage a pretrained model and fine-tune its Variational Auto Encoder (VAE) and retrain the Conditional Diffusion Model (CDM) components for our specific task.

\noindent\textbf{VAE Fine-tuning.} The standard VAE, when pretrained on general datasets, may not be sensitive enough to capture the fine-grained spatial and glyph features of scaled text instances. As highlighted by DiffUTE~\cite{DiffUTE} and also confirmed by our ablation experiments, fine-tuning the VAE is essential when working with text images. Therefore, we fine-tune it to enhance the capacity for accurate representation of text regions in a fair, local manner.   
The VAE consists of an encoder $ E_{\theta} $ and a decoder $ D_{\theta} $, and we let $ V_{\theta} $ represent the reconstruction after the encoding and decoding process. Following the pre-defined $ I_s $, we minimize the following loss function:
\begin{equation}
\mathcal{L}_{\text{TextSSR-VAE}} = \| V_{\theta}(I_s) - I_s \|_2^2 \label{vae}
\end{equation}

\begin{table*}[t!]\footnotesize
\centering
\begin{tabular}{c|cc|c|cc|c|cc|c}
\toprule

\multirow{2}{*}{\textbf{Method}} & \multicolumn{3}{c|}{\textbf{IC13}~\cite{ic13}} & \multicolumn{3}{c|}{\textbf{IC15}~\cite{ic15}} & \multicolumn{3}{c}{\textbf{ShopSign}~\cite{shopsign}} \\ 
\cmidrule(lr{0pt}){2-10}
                        & \multicolumn{1}{c}{SeqAcc(\%)$\uparrow$} & NED $\uparrow$  & FID-R$\downarrow$
                        & \multicolumn{1}{c}{SeqAcc(\%)$\uparrow$} & NED $\uparrow$  & FID-R$\downarrow$
                        & \multicolumn{1}{c}{SeqAcc(\%)$\uparrow$} & NED $\uparrow$  & FID-R$\downarrow$
                        \\ \midrule
TextDiffuser~\cite{textdiffuser} & 25.08 & 0.4779 & 103.74 & 0.82 & 0.0551 & 120.66 & 0.64 & 0.0230 & 136.56 \\
GlyphControl~\cite{glyphcontrol} & 0.55 & 0.0472 & 142.91 & 0 & 0.0217 & 152.37 & 0 & 0.0067 & 140.91 \\
UDiffText~\cite{udifftext} & 37.51 & 0.5892 & 67.82 & 2.55 & 0.1206 & 103.43 & --- & --- & --- \\
AnyText~\cite{anytext} & 55.07 & 0.7153 & 75.46 & 5.97 & 0.2186 & 76.03 & 33.58 & 0.4602 & 89.25 \\
TextDiffuser-2~\cite{textdiffuser2} & 26.39 & 0.5342 & 68.31 & 0.87 & 0.0838 & 82.93 & 0.82 & 0.0411 & 101.46 \\
\midrule
TextSSR~Pre-training & 75.35 & 0.9007 & \textbf{51.75} & 50.46 & 0.7361 & \textbf{34.69} & 30.28 & 0.5075 & 54.24 \\
TextSSR~Fine-tuning & \textbf{85.17} & \textbf{0.9402} & 53.28 & \textbf{62.64} & \textbf{0.8087} & 36.30 & \textbf{52.24} & \textbf{0.7053} & \textbf{52.84} \\
\midrule
Real & 96.51 & 0.9895 & 0 & 84.88 & 0.9486 & 0 & 74.84 & 0.8699 & 0 \\
\bottomrule
\end{tabular}
\caption{Comparing TextSSR (differences between the pre-training and fine-tuning versions are detailed in Appendix~\ref{sec:train_details}. We use the fine-tuned version, abbreviated as TextSSR, in the rest part of the paper.) and existing synthesis methods on regular text (IC13), irregular text (IC15), and multi-language text (ShopSign) datasets, alongside results from real datasets (Real). SeqAcc and NED denote the recognition accuracy, editing distance, respectively. FID-R score means FID on the text region.}
\label{tab:exp_acc}
\end{table*}

\noindent\textbf{CDM Retraining.}
To mask the actual text instance region while obtaining surrounding contextual information, we apply $I_M$ to $I_s$, resulting in $I_m$, through the operation $I_m = I_M \cdot I_s$.
$I_m$, $I_G$, and $I_s$ are fed into the frozen VAE, generating latent representations $Z_m$, $Z_G$, and $Z_s$. Additionally, $Z_s$ is further processed with added noise to produce $Z_T$. Moreover, we downsample $I_M$ to match their dimensions (e.g., in SD~2.1~\cite{Rombach_2022_CVPR_ldm}, height = width = 16), transforming it into $Z_M$.

These latent features, $Z_M$, $Z_T$, $Z_m$, and $Z_G$, are concatenated and then passed through a Conv2D layer to match the channel dimension required by the U-Net (e.g., in SD~2.1, channel = 4). Meanwhile, the Glyph Encoder (see Fig.~\ref{fig:model}(a)) that produces $F_g$ is also integrated into the CDM to extract features from $I_g$.
Finally, the CDM processes these inputs and produces the denoised latent representation $Z_{T-1}$, progressively refining the noisy input to recover the clean latent space. Assuming $\epsilon$ represents the original added noise and $\epsilon_\theta$ denotes the entire denoising process that outputs the predicted noise, the modified CDM is trained by minimizing the following objective:
\begin{equation}
\mathcal{L}_\text{TextSSR-CDM} = \| \epsilon - \epsilon_\theta(z_t, t, Z_M, Z_m, Z_G, I_g) \|_2^2
\label{cdm}
\end{equation}

\subsection{Anagram-based Data Synthesis}
\label{sec:3-4}

As mentioned above, our pipeline uses text region and content detected by an OCR engine as inputs. The recognized content is regarded as pseudo-label of the text. Our generation produces a synthesized text instance on the detected region for each pseudo-label rather than the groundtruth, which is typically unknown. Therefore, errors in OCR would not affect the generation. To expand the generation in quantity, we propose an anagram-based data expansion strategy. Unlike previous approaches that arbitrarily edit text~\cite{textdiffuser, anytext, textdiffuser2} or substitute equal-length strings~\cite{udifftext, DreamText}, we manipulate internal character order to maintain better region-content coherence. Meanwhile, the diffusion process adaptively handles the varying occupation requirement of different characters, e.g., `o' typically is wider than `l' in width. Therefore, our TextSSR leads to a more flexible generation. By changing the internal character order only, the region remain unchanged while still generating many samples with different character arrangements.

Our model's character-level rendering capability enables correct generation of text from this permutation-based operation. We define the total number of possible permutations for a word of length $l$ as:
\begin{equation}
P(l) = l!
\label{anagram}
\end{equation}

Note that some words may repeat due to character duplication; they still represent different samples, as the diffusion model is inherently random and there are also variations in cropped regions.

\section{Experiments}

\subsection{TextSSR Data Evaluation}

We first evaluate both accuracy and realism of the generation text, as well as the scaling capabilities of TextSSR data in small-scale. Then, we construct large-scale training data and assess its effectiveness at larger scales. Fig.~\ref{fig:viz} visualizes the examples of TextSSR compared to other methods.

\begin{figure*}
    \centering
    \includegraphics[width=0.8\linewidth]{./imgs/viz.pdf}
    \caption{Examples synthesized by TextSSR and other existing open-source methods (UDiffText only supports English text for its limited text encoder). TextSSR consistently produces accurate and realistic text instances. Please refer to Appendix~\ref{sec:viz_analysis} for more analysis. }
    \label{fig:viz}
\end{figure*}

\noindent\textbf{Accuracy Evaluation.} Evaluating the accuracy of generated scene text is challenging, as manually verifying alignment with target text is costly. Previous benchmarks typically employ an STR model to automatically assess text consistency. We also follow this scheme. To mitigate potential OCR recognition errors and more effectively reflect the accuracy and difficulty level of our synthetic data, we choose SVTRv2~\cite{du2024svtrv2}, a SOTA STR model upgraded from SVTR~\cite{du2022svtr}, as the recognition model.

Tab.~\ref{tab:exp_acc} gives the accuracy assessment results across different datasets, where \textbf{IC13} denotes all the methods using text instances from the IC13 dataset~\cite{ic13} to initialize their generation, while others are defined similarly. As can be seen, a large portion of text instances generated by TextSSR are correctly recognized. For more irregular and difficult \textbf{IC15}~\cite{ic15}, and less linguistically \textbf{ShopSign}~\cite{shopsign} datasets, despite accuracy declines observed, TextSSR still outperforms the others by significant margins. The results provide strong evidence of TextSSR's correctness. 

Comparing with AnyText~\cite{anytext}, the current SOTA, TextSSR outperforms it by 30.1\% and 56.7\% in accuracy on IC13 and IC15, respectively. This is mainly because TextSSR operates generation at the instance level, while AnyText performs on the full image level. Note that methods based on full-image synthesis struggle significantly in complex, challenging scenarios, e.g., IC15. This dataset includes many text instances affected by blur, low-resolution, perspective distortion, and curvature—challenges that previous methods do not address effectively. In contrast, TextSSR benefits from a region-centered generation pipeline. It excels at handling challenging text instances and maintains reasonable synthesis accuracy. Moreover, the condition for TextSSR is glyph shapes, which can be extended to other languages like Chinese. The results on ShopSign prove TextSSR's multilingual capability. It outperforms the purportedly multilingual-focused AnyText by 18.7\%. More details on accuracy evaluation are provided in Appendix~\ref{sec:acc_details}.

\begin{table}[t]\footnotesize
\centering

\begin{tabular}{ccccc}

\toprule
\multirow{2}{*}{\textbf{Type}} & \multirow{2}{*}{\textbf{Method}} & \multicolumn{3}{c}{\textbf{SeqAcc(\%)$\uparrow$}}  \\ 
\cmidrule(lr{0pt}){3-5} & & Regular & Irregular & Avg
                        \\  
\midrule
\multirow{3}{*}{\textbf{Rendering}} & SynthText~\cite{st} & 48.55 & 26.08 & 39.74 \\ 
& VISD~\cite{visd} & 38.02 & 26.88 & 33.65 \\ 
& UnrealText~\cite{unrealtext} & 32.06 & 19.85 & 27.28 \\ 
\midrule
\multirow{6}{*}{\textbf{Diffusion}} & TextDiffuser~\cite{textdiffuser} & 30.08 & 8.79 & 21.74 \\ 
& GlyphControl~\cite{glyphcontrol} & 40.80 & 13.21 & 29.99 \\ 
& AnyText~\cite{anytext} & 49.80 & 19.62 & 37.97 \\ 
& TextDiffuser-2~\cite{textdiffuser2} & 40.26 & 10.48 & 28.59 \\ 
& SceneVTG~\cite{SceneVTG} & 54.97 & 35.50 & 47.34 \\ 
& TextSSR (Ours) & \textbf{59.76} & \textbf{35.71} & \textbf{50.33} \\
\midrule
\multirow{2}{*}{\textbf{Real}} & Real-L~\cite{baek2021whatif} & \underline{57.97} & 37.66 & 50.01 \\ 
& TextOCR~\cite{singh2021textocr} & 57.83 & \underline{41.56} & \underline{54.05} \\ 
\bottomrule
\end{tabular}
\caption{Comparison of different data synthetic methods, where CRNN~\cite{crnn} trained on different synthetic data and real data of 30k is employed to obtain the results. ``Regular" denotes the average on IIIT~\cite{iiit}, SVT~\cite{svt}, and IC13~\cite{ic13} datasets, ``Irregular" represents the average on IC15~\cite{ic15}, SVTP~\cite{svtp}, and CUTE~\cite{cute} datasets, while ``Avg" is the average of the six datasets.}
\label{tab:exp_real}
\end{table}

\begin{table*}[t]\footnotesize
\centering
\setlength{\tabcolsep}{3pt}{
\resizebox{0.94\textwidth}{!}{
\begin{tabular}{cccccccccc|c}
\toprule
\multirow{2}{*}{\textbf{Model}} & \multirow{2}{*}{\textbf{Dataset}} & \multirow{2}{*}{\textbf{Volume}} & \multicolumn{8}{c}{ \textbf{SeqAcc(\%)$\uparrow$}}   \\ 
\cmidrule(lr{0pt}){4-11}
&  & &  IIIT5k~\cite{iiit} & SVT~\cite{svt} & IC13~\cite{ic13} & IC15~\cite{ic15} & SVTP~\cite{svtp} & CUTE80~\cite{cute} & Avg & Contextless~\cite{jiang2023revisiting}
                        \\ \midrule
\multirow{8}{*}{CRNN~\cite{crnn}} 
& ST & 4m & 90.13 & 82.84 & 90.90 & 72.45 & 72.09 & 82.99 & 81.90 & 33.50  \\ 
& ST & 6.98m & 90.80 & 83.00 & 91.83 & 72.34 & 73.80 & 81.60 & 82.06 & 32.61  \\ 
& ST~+~SynthAdd & 4m~+~1.05m & 90.43 & 83.93 & 91.13 & 73.55 & 72.71 & 80.21 & 81.99 & 34.53  \\
&  ST~+~TextSSR & 4m~+~1.05m & 92.13 & 87.02 & 93.35 & 77.42 & 77.67 & \underline{84.37} & 85.33 & 55.97 \\
&  ST~+~TextSSR & 4m~+~3.55m & \underline{93.57} & \underline{89.18} & \underline{94.28} & 78.58 & 78.45 & 82.99 & \underline{86.17} & \underline{57.89} \\
\cline{2-11} 
\\[-1em]
& ST~+~TextOCR & 4m~+~1.05m & 93.13 & 88.10 & 92.53 & \underline{79.68} & \textbf{79.22} & 83.68 & 86.06 & 44.93 \\
& ST~+~TextOCR & \multirow{2}{*}{5.05m~+~3.55m} & \textbf{94.57} & \textbf{89.34} & \textbf{94.40} & \textbf{80.23} & \textbf{79.22} & \textbf{84.72} & \textbf{87.08} & \textbf{58.15} \\
   & ~+~TextSSR &  & \textcolor{blue}{+1.44} & \textcolor{blue}{+1.24} & \textcolor{blue}{+1.87} & \textcolor{blue}{+0.55} & +0.00 & \textcolor{blue}{+1.04} & \textcolor{blue}{+1.02} & \textcolor{blue}{+13.22} \\

\midrule
\multirow{8}{*}{MAERec~\cite{jiang2023revisiting}} 
& ST & 4m & 96.00 & 91.96 & 95.80 & 83.88 & 85.74 & 89.58 & 90.49 & 59.43  \\ 
& ST & 6.98m & 96.90 & 92.89 & 95.80 & 84.70 & 86.82 & 92.36 & 91.58 & 61.62  \\ 
& ST~+~SynthAdd & 4m~+~1.05m & 95.97 & 93.20 & 95.80 & 85.04 & 86.98 & 90.28 & 91.21 & 60.46  \\
&  ST~+~TextSSR & 4m~+~1.05m & 97.53 & 93.82 & 96.85 & 86.75 & 89.46 & 94.10 & 93.08 & 76.77 \\
&  ST~+~TextSSR & 4m~+~3.55m & \underline{98.13} & 94.74 & \textbf{97.32} & 87.85 & 89.61 & 94.10 & 93.63 & \underline{78.17} \\
\cline{2-11} 
\\[-1em]
& ST~+~TextOCR & 4m~+~1.05m & 98.10 & \underline{95.52} & 97.20 & \underline{89.34} & \underline{90.39} & \underline{96.87} & \underline{94.57} & 74.33 \\
& ST~+~TextOCR & \multirow{2}{*}{5.05m~+~3.55m} & \textbf{98.20} & \textbf{96.75} & \underline{97.20} & \textbf{89.95} & \textbf{92.71} & \textbf{97.92} & \textbf{95.46} & \textbf{82.03} \\
& ~+~TextSSR &  & \textcolor{blue}{+0.10} & \textcolor{blue}{+1.23} & +0.00 & \textcolor{blue}{+0.61} & \textcolor{blue}{+2.32} & \textcolor{blue}{+1.05} & \textcolor{blue}{+0.89} & \textcolor{blue}{+7.70} \\

\bottomrule
\end{tabular}}
}
\caption{Performance of models trained on different data combinations, where CRNN and MAERec are selected as model representatives. \textbf{Bold} and \underline{underline} indicate the highest and second highest values, respectively.}
\label{tab:exp_usability}
\end{table*}

\noindent\textbf{Realism Evaluation.} Following the validation presented in SceneVTG~\cite{SceneVTG}, we generate a fixed amount (30k) of data to train an identical model under fair settings and then test it on common STR benchmarks~\cite{iiit, svt, ic13, ic15, svtp, cute}. This approach avoids the need for separate verification of text correctness or synthesis quality. If the generated text is inaccurate or unrealistic, this will be reflected in poor recognition performance during testing. 

As shown in Tab.~\ref{tab:exp_real}, for the compared methods, their average SeqAcc is below 40\% except for SceneVTG~\cite{SceneVTG}. The result is closely linked to their low generation accuracy. Note that rendering-based methods generally perform better than traditional diffusion-based methods in irregular benchmarks, highlighting the challenge of generating highly controllable text images. SceneVTG benefits from its two-stage pipeline of text erasure and then rendering, it reaches a SeqAcc of 47.3\%. Nevertheless, TextSSR, featured by its glyph-based condition injection and anagram-based generation, surpasses SceneVTG by nearly 3\%. Moreover, the gap between models trained on 30k TextSSR and real-world TextOCR is only 3.7\%, which comes from irregular text. These results convincingly demonstrate the realism of TextSSR data, especially in regular text synthesizing. However, further efforts are still needed to improve the synthesis of irregular text.

\noindent\textbf{Scalability Evaluation}. TextSSR data can be easily scaled up to a large quantity. It is necessary to assess whether the increase in data volume can translate into performance improvements. Therefore, we use the model generated by the 30k data in the realism evaluation as baseline (TextSSR $\times$1), and re-train CRNN under the same experimental setup but with data doubled every time. Results of this scalability evaluation are presented in Tab.~\ref{tab:ablation data}. When scaled up to $\times$4 and $\times$8, accuracy improvements of 4.7\% and 14.1\% are observed, respectively. The result indicates that increasing the data volume indeed contributes to better STR models. Note that models in the two scenarios also surpass the model trained on 30k real-world data by 1.0\% and 10.4\%, respectively. It is observed that TextSSR $\times$8 also presents a much better result than real data model in irregular text, highlighting that the challenge of irregularity can be overcome to some extent by increasing the data volume. Meanwhile, we also conducted a comparison experiment similar to \cite{udifftext}, i.e., substituting text instances with equal-length strings (TextSSR EL), where the trained model reported quite worse results. This again demonstrates the effectiveness of our anagram-based generation pipeline.

\begin{table}[t]\small
\centering

\setlength{\tabcolsep}{4pt}{
\begin{tabular}{lcccc}

\toprule
\multirow{2}{*}{\textbf{Method}} & \multirow{2}{*}{\textbf{Volume}} & \multicolumn{3}{c}{\textbf{SeqAcc(\%)$\uparrow$}}  \\ 
\cmidrule(lr{0pt}){3-5}
&  & Regular & Irregular & Avg
                        \\ \midrule
                        
TextSSR~$\times$1 & 30k & 59.76 & 35.71 & 50.33 \\
TextSSR~EL & 30k & 24.58 & 10.23 & 18.95 \\
\midrule
TextSSR~$\times$2 & 60k & 60.49 & 34.49 & 50.29 \\
TextSSR~$\times$4 & 120k & 65.17 & 39.34 & 55.03 \\
TextSSR~$\times$8 & 240k & \textbf{73.96} & \textbf{49.69} & \textbf{64.44} \\
\midrule
TextOCR & 30k & 57.83 & 41.56 & 54.05 \\ 
\bottomrule

\end{tabular}}
\caption{TextSSR scalability evaluation. $\times n$ refers to perform the anagram-based rendering $n$ times. EL denotes equal length edit.}
\label{tab:ablation data}
\end{table}

\noindent\textbf{Usability Assessment}. We further assess the effectiveness of our synthesis method in supporting the large-scale training of STR models. To ensure the quality of data, we apply a quality-screened step to every sample generated by TextSSR. We still use SVTRv2~\cite{du2024svtrv2} to recognize the sample and compare the recognized content with its generation label. The sample is reserved if they are consistent, otherwise we discard it. The results on different training data combinations are shown in Tab.~\ref{tab:exp_usability}. The observations can be summarized as follows:
First, the rendering-based ST dataset reaches a saturation point in terms of accuracy, where 6.98m samples only show marginal improvement (0.16\% on CRNN) over the 4m dataset on the six common benchmarks. This suggests that we have to move towards more realistic data synthesis.
Second, when incorporating 1.05m our synthetic dataset, the two STR models (CRNN and MAERec) gain improvements of 3.4\% and 2.6\% over the 4m ST dataset on the six common benchmarks, respectively. They are only 0.7\% and 1.5\% lower than ST combined with equivalent size real-world TextOCR. This suggests that TextSSR data is approaching real-world data in quality.
Third, when further increasing TextSSR data to 3.55m, it is seen that the purely synthetic training data (ST~+~3.55m TextSSR) gain performance more approaching the synthetic-real mixed training data (ST~+~1.05m TextOCR). Moreover, when combining the 3.55m synthetic data with this mixed data, the trained model can report further performance improvements, obtaining 1.02\% on CRNN and 0.89\% on MAERec. This implies that TextSSR data is a valuable supplement to existing training data. Using it individually or collectively both can benefit STR model training. As a result, we mark this 3.55m dataset as TextSSR-F, an accurate and realistic large-scale synthetic dataset constructed by our TextSSR, and make it publicly available.

Since our anagram-based synthesis generates a large number of contextless words, and contextless word generation is also discussed in SynthAdd~\cite{li2019show}, we  conduct a comparison between SynthAdd and our TextSSR. As shown in Tab.~\ref{tab:exp_usability}, there are clear performance margins (3.3\% on CRNN and 1.9\% on MAERec) between the model trained on traditional SynthAdd and TextSSR. This also indicates the superiority of TextSSR as a STR training data synthetic method.

\begin{table}[h]\footnotesize 
  \centering
 \setlength{\tabcolsep}{2pt}{
    \begin{tabular}{cccc|cccc}
      \toprule
       \begin{tabular}[c]{@{}c@{}}region-\\ centric\end{tabular}  & \begin{tabular}[c]{@{}c@{}}char-\\ glyph\end{tabular} & \begin{tabular}[c]{@{}c@{}}char-\\ position\end{tabular} & \begin{tabular}[c]{@{}c@{}}fine-tune\\ VAE\end{tabular}  & SeqAcc (\%) $\uparrow$ & NED $\uparrow$ & FID-R $\downarrow$ \\
        \midrule
         &  &  &   & 20.28 & 0.3444 & 102.89 \\
        \ding{51} &  &  &  & 73.50 & 0.8764 & 45.89 \\
        \ding{51} & \ding{51} &  &  &  74.15 & 0.8852 & 46.65 \\
        \ding{51} & \ding{51} & \ding{51} &  &  80.26 & 0.9133 & \textbf{45.83} \\
        \ding{51} & \ding{51} &  & \ding{51} &  77.21 & 0.9043 & 50.34 \\
        \ding{51} &  &  & \ding{51} & 76.23 & 0.8991 & 51.94 \\
         & \ding{51} & \ding{51} & \ding{51} &  47.55 & 0.6211 & 62.52 \\
        \ding{51} & \ding{51} & \ding{51} & \ding{51} &  \textbf{82.99} & \textbf{0.9308} & 50.76 \\
      \bottomrule
    \end{tabular}}
    \caption{Ablation study of TextSSR components using the TextSSR pre-training model in Tab.~\ref{tab:exp_acc} on the IC13 dataset.}
    \label{table:ablation components}
\end{table}

\subsection{Ablation Study}

To assess the proposed components, we start with the TextSSR pre-training model in Tab.~\ref{tab:exp_acc} (to simplify, trained for only 5k steps), where all components are considered. Then, we stepwisely remove one or some of the components to assess their respective necessity and effectiveness. From the results on Tab.~\ref{table:ablation components}, we can see that the removal of each one brings a clear decline in performance, while the absence of  ``region-centric" processing can lead to a catastrophic drop of 35.4\% in accuracy. Additionally, fine-tuning the VAE offers a trade-off, slightly sacrificing visual quality for an improvement in accuracy. This is because real-world scene text also confronts poor visual quality challenge, simulating this to some extent can provide valuable samples for model training.

\begin{figure}[h]
    \centering
    \includegraphics[width=0.8\linewidth]{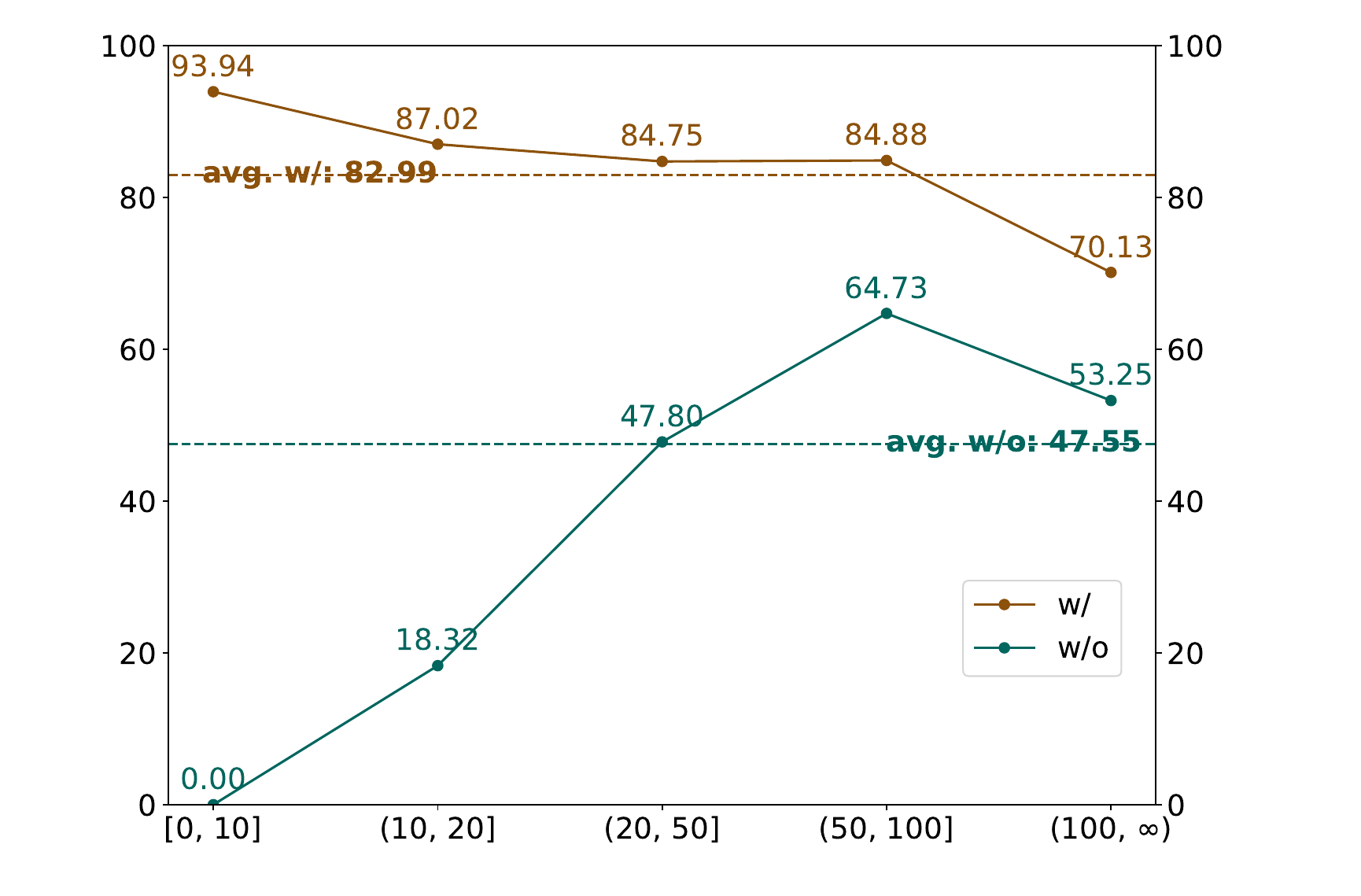}
    \caption{Performance of TextSSR fine-tuning model with and without region-centric processing across text regions of different sizes (in pixels).}
    \label{fig:plot}
\end{figure}

We perform another experiment to verify the effectiveness of the region-centric processing. The results in Fig.~\ref{fig:plot} show that without this processing, TextSSR has to synthesize text instance on the whole image. It exhibits a performance trend that first rises and then falls as the size increases. When the region is small, the entire image is scaled severely, resulting in more blurred text regions and insufficient pixels for text rendering. As the region size increases, larger regions and more pixels are allowed, therefore performance increase. When a too large region given, it causes the side effect that constraints the shape of the generated text, and performance decline. In contrast, incorporating region-centric processing allows TextSSR to achieve superior and consistent performance across all sizes.

\begin{figure}[h]
    \centering
    \includegraphics[width=0.9\linewidth]{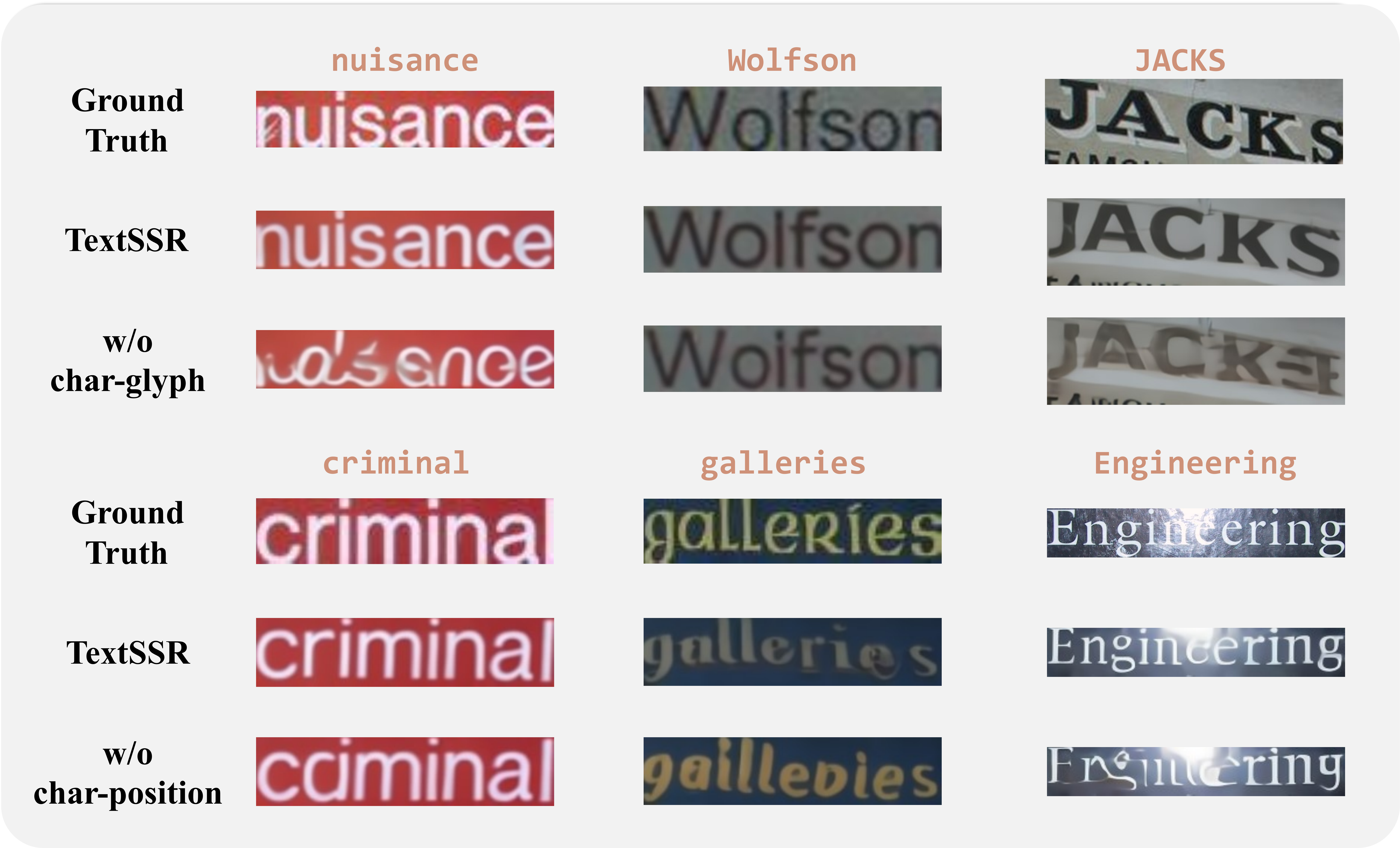}
    \caption{Visualizations of TextSSR with and without character-level prior. More examples are provided in Appendix~\ref{sec:abla_viz} }
    \label{fig:wocharpg}
\end{figure}

In Fig.~\ref{fig:wocharpg}, we also illustratively ablate the character-level position and glyph components. The results indicate that omitting either component leads to issues such as character deformation, incorrect characters, and duplication errors.

\section{Conclusion}

We have proposed TextSSR to provide STR model training with high-quality synthetic data. It employs the glyph of existing text and surrounding context as the prompt of the diffusion-based synthesizing model, and combines character-level constraints and permutations. TextSSR successfully synthesizes accurate and realistic text instances at a large scale. The TextSSR-F dataset with 3.55 million diverse and realistic instances are constructed. We have conducted extensive experiments to assess TextSSR-F. STR models trained solely on TextSSR-F, and on the combination of TextSSR-F and real-world data both show performance improvements. While the effectiveness of TextSSR has been basically confirmed, for irregular text, there is still an accuracy gap between models trained on TextSSR-F and real-world data. Thus, future work includes the exploration of more controllable ways to further improve the synthesizing quality, especially for challenging instances such as curved and multi-oriented text. We are also interested in extending TextSSR to generate large-scale data in different languages like Chinese.

\noindent\textbf{Acknowledgement}
This work was supported by the National Natural Science Foundation of China (Nos. 32341012, 62172103).

{
    \small
    \bibliographystyle{ieeenat_fullname}

}

\clearpage
\setcounter{page}{1}

\maketitlesupplementary

\section{More Implementation Details}
\label{sec:implementation}

\subsection{Model specific settings}

In the setup of Sec.~\ref{sec:3-2}, we consider the specific characteristics of scene text: \( L \) is set to 25, and \( P_M \) is configured to 128, assuming a background color of 255. It is worth noting that our method is not limited to this configuration. This configuration theoretically enables the generation of text up to 255 characters in length, allowing for background color selection from 255 and character values from 0 to 254.  Each character image is set to a 64$\times$64 square, resulting in a character glyph image of size 25$\times$64$\times$64. The deformed ViT is configured with a patch size of 8, generating a latent feature vector of size 65$\times$1024, where 1024 represents the dimensionality of the control information required by the CDM. 

After passing through the VAE in Sec.~\ref{sec:3-3}, the dimensions of outputs are uniformly [4, 16, 16], while $Z_M$ is formatted as [1, 16, 16]. These are concatenated to form [13, 16, 16]. Therefore, the Conv2d Layer has an input dimension of 13 and an output dimension of 4, producing an output of [4, 16, 16] to match the original input dimensions of the U-Net.

To meet the rendering requirements for visible characters in the majority of languages, we utilize  to employ Puhui Font, an open-source and commercially-free font tool. It adheres to the latest Chinese national standard, GB18030-2022, and supports 178 languages.

\subsection{More Details for Datasets}
\label{sec:dataset}
We will further elaborate on the data processing related to training and generation.

\textbf{Training Data.}  To train our generative model, we utilize the large-scale multilingual text image dataset, AnyWord-3M~\cite{anytext}. It contains real annotated text boxes and text contents, designed for scene text detection and recognition tasks, which we collectively call AnyWord-Scene. This collection includes a range of popular datasets such as ArT~\cite{ArT}, COCO-Text~\cite{veit2016coco}, RCTW~\cite{shi2017icdar2017rctw}, LSVT~\cite{sun2019icdarlsvt}, MLT~\cite{nayef2019icdar2019mlt}, MTWI~\cite{mtwi}, and ReCTS~\cite{liu2019icdarrects}. In addition, two larger datasets are included: AnyWord-Wukong 
~\cite{gu2022wukong} and AnyWord-Laion 
~\cite{schuhmann2021laion}, which provide a large collection of images with bounding boxes and text content obtained by the PP-OCRv3 
~\cite{li2022pp} detection and recognition model. We filter out anomalous images with pure white backgrounds from the AnyWord-3M dataset. Ensuring that when cropping local images, we minimize the inclusion of white borders, thereby reflecting real-world conditions. The processed AnyWord-Wukong and AnyWord-Laion datasets together contain a total of 3,430,412 complete images, from which we crop 14,856,392 local regions containing text instances for the first training stage. In the second training stage, we utilize 78,395 full images from processed AnyWord-Scene, cropping 201,599 text regions from these.

\textbf{Computational Overhead and Runtime Efficiency.} The training times for our three stages (VAE fine-tuning, UNet pretraining, and UNet fine-tuning) are 192, 400, and 200 GPU-hours, respectively. For inference, using a single RTX-3090 GPU with diffusion\_steps=20 and batch\_size=32, 5k batches take 26 hours, averaging 0.59 seconds per image.

\begin{figure}
    \centering
    \includegraphics[width=1\linewidth]{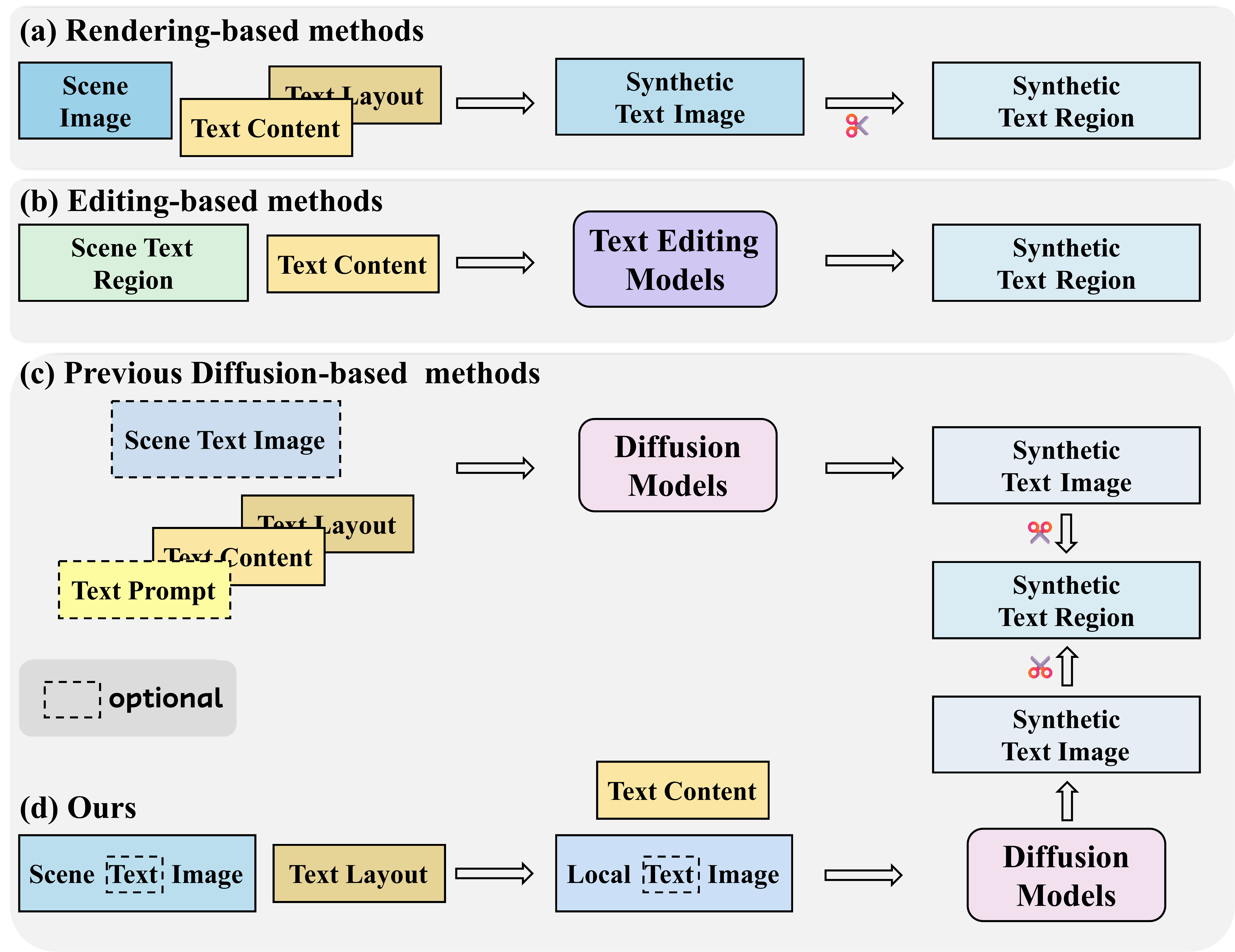}
    \caption{The pipeline comparison between TextSSR and previous methods. In (c), ``optional" indicates that either or both options should be provided. And in (d), it means that the base image can be used with or without text.}
    \label{fig:compare}
\end{figure}

\begin{table*}[t]
\footnotesize
\centering
\resizebox{\textwidth}{!}{ 
\begin{tabular}{c|ccc|ccc|ccc|ccc}
\toprule

\multirow{2}{*}{\textbf{Method}} & \multicolumn{3}{c|}{\textbf{French}} & \multicolumn{3}{c|}{\textbf{German}} & \multicolumn{3}{c|}{\textbf{Japanese}} & \multicolumn{3}{c}{\textbf{Traditional Chinese}} \\ 
\cmidrule(lr{0pt}){2-13}
                        & SeqAcc(\%)$\uparrow$ & NED$\uparrow$  & FID-R$\downarrow$
                        & SeqAcc(\%)$\uparrow$ & NED$\uparrow$  & FID-R$\downarrow$
                        & SeqAcc(\%)$\uparrow$ & NED$\uparrow$  & FID-R$\downarrow$
                        & SeqAcc(\%)$\uparrow$ & NED$\uparrow$  & FID-R$\downarrow$ \\ \midrule
TextDiffuser   & 12 & 0.2953 & 231.47 & 5  & 0.1707 & 240.55 & 0 & 0.0014 & 280.95 & 0  & 0.0011 & 259.46 \\
GlyphControl   & 0  & 0.0188 & 268.63 & 0  & 0.0264 & 277.10 & 0 & 0.0000 & 276.20 & 0  & 0.0011 & 261.79 \\
AnyText        & 24 & 0.4508 & 181.22 & 16 & 0.3276 & 198.34 & 3 & 0.1696 & 215.89 & 21 & 0.3557 & 208.85 \\
TextDiffuser2  & 9  & 0.2891 & 169.56 & 6  & 0.2178 & 208.04 & 0 & 0.0015 & 230.02 & 0  & 0.0080 & 190.93 \\
\midrule
TextSSR        & \textbf{68} & \textbf{0.8024} & \textbf{107.78} & \textbf{75} & \textbf{0.8734} & \textbf{102.86} & \textbf{24} & \textbf{0.5545} & \textbf{127.87} & \textbf{55} & \textbf{0.7304} & \textbf{112.81} \\
\midrule
Real           & 94 & 0.9858 & 0 & 94 & 0.9810 & 0 & 83 & 0.9562 & 0 & 98 & 0.9967 & 0 \\
\bottomrule
\end{tabular}
}
\caption{Quantitative comparison of multilingual text image generation methods. For each language, 100 images are used for test.}
\label{tab:exp_acc_mlt}
\end{table*}

\textbf{Data for Accuracy Evaluation.}  Meanwhile, in Accuracy Evaluation we employ datasets that the model has not previously encountered. They represent a range of image difficulty levels and cover multiple languages. Specifically, we use the following datasets for evaluation:

\begin{itemize}
\item \textbf{IC13}~\cite{ic13}: This benchmark is designed for relatively regular text detection and recognition. We use 233 full images and 917 cropped text images from the test set for evaluation.
\item \textbf{IC15}~\cite{ic15}: This dataset contains more challenging real-world scene text, derived from incidental scene captures where the text was not the primary focus. We employ 500 full images and 2,077 cropped text images for evaluation.
\item \textbf{Shopsign}~\cite{shopsign}: This dataset consists of Chinese scene text, primarily from shop signs. We selecte 183 full images and 932 cropped text images from this dataset.
\end{itemize}

\textbf{Base Data for Scalable Generation.} We utilize the TextOCR~\cite{singh2021textocr}, a large-scale scene text dataset including 25,119 images, as the base images for synthetic data generation. To mimic realistic conditions where unlabeled data is abundant, we use pseudo-labels generated by the PP-OCRv3 model, thus simulating a scenario without human annotation. 


\textbf{Generation of TextSSR-F.} After the previous step, we obtain 188,526 text regions annotated by the PP-OCRv3~\cite{li2022pp} model. Based on the anagram method (described in Section~\ref{sec:3-4}), we expand the data and filter with the SVTRv2~\cite{du2024svtrv2} model, yielding a final dataset of 3,551,396 fully usable text instances, referred to as TextSSR-F.

\textbf{Quality Filtering Bias} Our filtering process employs a double-check mechanism: given the label the generation model attempts to generate, and SVTRv2 is used to verify that the recognized text matches the label. An error occurs only if both the generation model and SVTRv2 fail simultaneously. This pipeline ensures the correctness of most TextSSR-F instances. To validate this, we randomly sample 300 instances from TextSSR-F and recruit three assessors each checking 100 instances. The average accuracy is 98.67\% (98\%, 98\%, 100\%).

\begin{figure}[ht] \centering
     \includegraphics[width=0.48\textwidth]{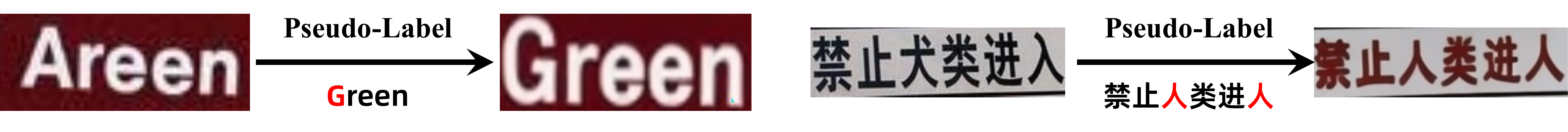}
    \caption{Examples of correct rendering despite incorrect English and Chinese pseudo-labels.} \label{fig:pseudo}
    \vspace{-1em}
\end{figure}

\textbf{Impact of Pseudo-Labeling}
When OCR pseudo-label errors occur, the generation process still attempts to follow the pseudo-labels (see examples in Fig.~\ref{fig:pseudo}), so the side impact is relatively limited. 

\vspace{-0.7em}

\subsection{Training and Evaluation Details}

\subsubsection{Training Details}
\label{sec:train_details}
We fine-tune our generative model using Stable Diffusion 2.1 (SD~2-1)~\cite{Rombach_2022_CVPR_ldm} on eight NVIDIA 3090 GPUs. First, we train the VAE on the full AnyWord dataset with a total batch size of 512 and 256x256 image patches for 150k steps. Then, we freeze the VAE and train the CDM in two stages: 50k steps on AnyWord-Wukong and AnyWord-Laion datasets to pre-train, followed by 25k steps on the AnyWord-Scene dataset to fine-tune, using a total batch size of 256. 

\subsubsection{Accuracy Evaluation Details}
\label{sec:acc_details}
For a fair comparison in our accuracy evaluation, we render all visible bounding boxes and contents annotated in the test datasets. In cases where certain models could not render longer texts or handle multiple text instances per image, we will restrict the input information to within their acceptable ranges, while padding the missing portions. Our model is also limited, with only the first 25 characters rendered for single-character features. For all models, the number of timesteps in the sampling process is set to 20. The evaluation code for generated results is based on the open-source evaluation scripts from AnyText~\cite{anytext} and UDiffText~\cite{udifftext}. Except for GlyphControl~\cite{glyphcontrol}, which requires additional image descriptions to function properly, the other methods only use their predefined text prompts.

\begin{figure*}[ht] \centering
    \includegraphics[width=\linewidth]{imgs/mlt.pdf}
    \caption{Visualization of synthesized multilingual examples.} \label{fig:mlt}
\end{figure*}

\subsubsection{Expanded Multilingual Evaluation}
We have added four languages (French, German, Japanese, and Traditional Chinese) and use the multilingual version of SVTRv2 for evaluation (see Tab.~\ref{tab:exp_acc_mlt}). We also provide illustrative examples in Fig.~\ref{fig:mlt} to validate the generalization to non-English languages. TextSSR generates correct instances while others mostly fail.

\subsubsection{Realism and Scalability Evaluation Details}
In the Realism and extended experiments, CRNN is trained~\cite{baek2021whatif} with a batch size of 64 on a single 3090 GPU for 10k steps. The data augmentation configurations preset in the codebase are utilized throughout the training process.

\subsubsection{Usability Assessment Details}
In the Usability experiments, we train two widely-used STR models—CRNN~\cite{crnn}, and MAERec~\cite{jiang2023revisiting}—on the generated data to assess the effectiveness of our synthetic data in enhancing STR performance. All models are trained using the OpenOCR framework, with a total batch size of 1024 on four 3090 GPUs for 20 epochs.

To vividly demonstrate that TextSSR significantly enhances the performance of STR models under challenging scenarios, we conduct a small-scale validation experiment. In this experiment, we limit the dataset size to 429k and employ an identical NRTR~\cite{sheng2019nrtr} model trained under the same configuration for comparison. The experimental results indicate that the model trained on TextSSR-F exhibits more realistic performance when dealing with challenging text conditions such as perspective distortion and blurring. We provide visual comparisons in Fig.~\ref{fig:rec_result}, showcasing TextSSR's superior performance in recognizing low-resolution and perspective-distorted text.

\begin{figure}[H] 
    \centering
    \includegraphics[width=0.48\textwidth]{imgs/rec_result.pdf}
    \caption{Visualization of recognition results on NRTR.} \label{fig:rec_result}
\end{figure}

\subsubsection{Ablation Study Details}
The ablation study can be considered a simplified version of the second stage of training, with all settings kept consistent except for the reduction of training steps to 5k. To align with the full-image inference process used in other methods, the image size is set to 512$\times$512, although training is conducted at a resolution of 256. The ``Char-Glyph" ablation experiment involves removing the condition from the CDM training, while the ``Char-Position" ablation renders all characters uniformly at a pixel value of 127. 

\begin{figure*}[h!]
    \centering
    \includegraphics[width=\linewidth]{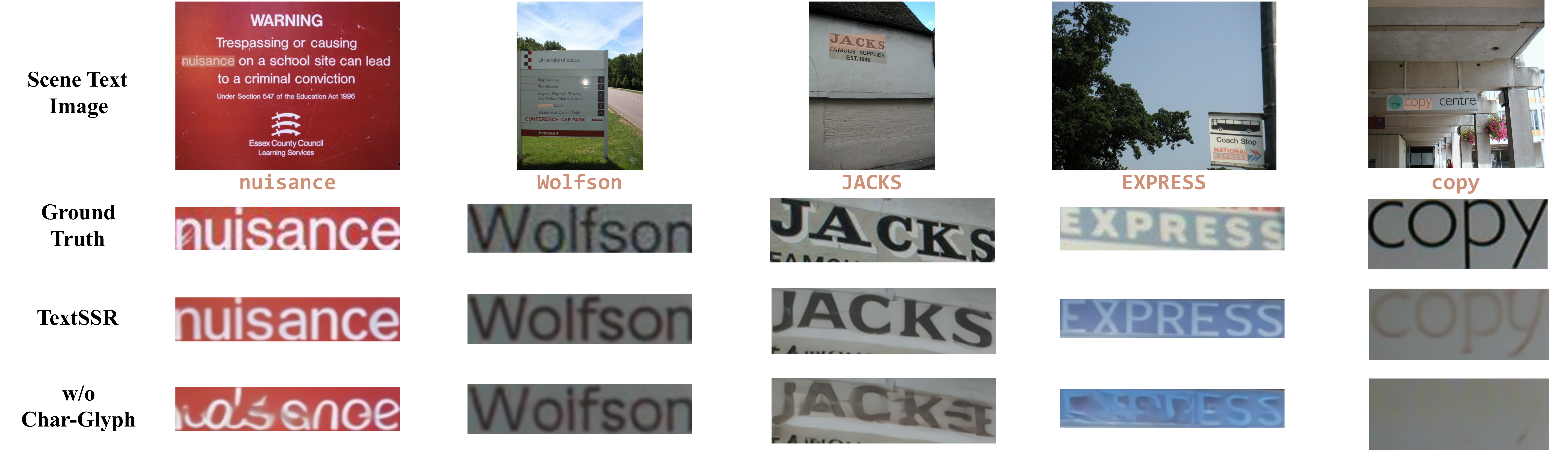}
    \caption{Visualization results of TextSSR with and without character-level glyph prior.}
    \label{fig:wocg}
\end{figure*}

\section{Visualization}

\subsection{Visualization Analysis}
\label{sec:viz_analysis}

Fig.~\ref{fig:viz} sequentially simulates various situations, including English text in a regular scene, text under challenging conditions, and Chinese text in a natural setting. TextSSR consistently generates accurate and high-quality visual text, demonstrating several powerful capabilities: (1) it can synthesize arbitrary text with standard glyphs from any language, as shown in examples of both Chinese and English; (2) it learns font style information from surrounding context, such as the font color in Sample 1, which is derived from the horizontal line below; (3) it synthesizes correct text even without strong background information, as illustrated in Sample 2, where the local image provides no usable information for imitation; and (4) it exhibits scale invariance, allowing for text synthesis in scenes of any size, with the three samples representing large, small, and medium text sizes, respectively.

\begin{figure}[h]
    \centering
    \includegraphics[width=1\linewidth]{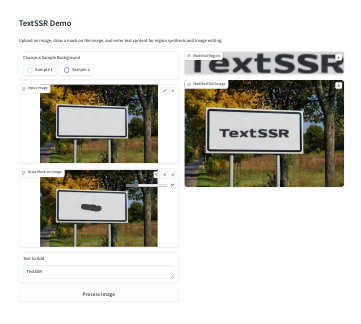}
    \caption{Users will select a scene image as the base, perform mask marking in the designated area, and then input the text content to be written. After processing, the desired text region and the edited original image will be obtained.}
    \label{fig:demo}
\end{figure}

\subsection{Function Demonstration Platform}

We have concretely implemented the inference process and build a demonstration demo.   To ensure that the user input matches the label format used during training, we recalculate text boxes that align with the input text location after the user applies the mask.   As shown in Fig.~\ref{fig:demo}, the text is roughly displayed within the user-specified area, though it does not follow the mask strictly.

\subsection{Ablation Visualization Results}
\label{sec:abla_viz}

\begin{figure*}[h!]
    \centering
    \includegraphics[width=\linewidth]{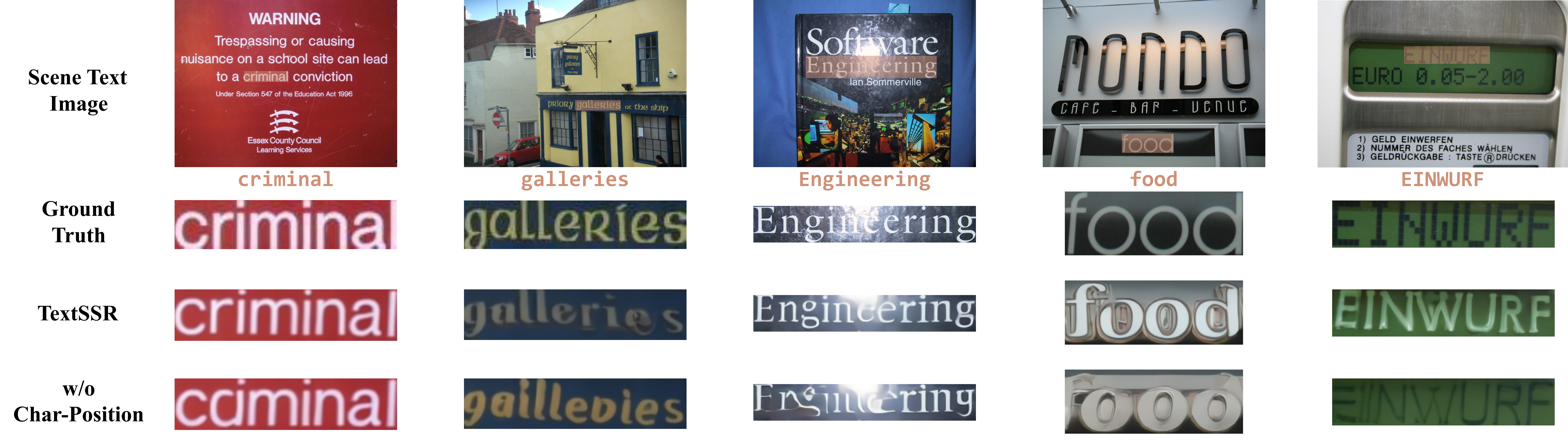}
    \caption{Visualization results of TextSSR with and without character-level position prior.}
    \label{fig:wocp}
\end{figure*}

Fig.~\ref{fig:wocp} and Fig.~\ref{fig:wocg} illustrate the impact of character-level position and glyph on the rendering results of TextSSR. The results indicate that omitting either component leads to issues such as character deformation, incorrect characters, and duplication errors in some cases, further supporting the findings of the ablation study.

\subsection{More Visualization Results}
To provide a detailed illustration of the synthesis process and effectiveness of TextSSR, Fig.~\ref{fig:more_viz} illustrates TextSSR's synthesis process, showing how it reconstructs local images from original regions and crops them to obtain final results. Comparisons with ground truth demonstrate TextSSR's strong synthesis capabilities across diverse scenarios, including regular text, low-resolution text, curved text, perspective text, multilingual text and multi-oriented text.

\subsection{Failure Cases}

\begin{figure*}[t]
    \centering
    \includegraphics[width=\linewidth]{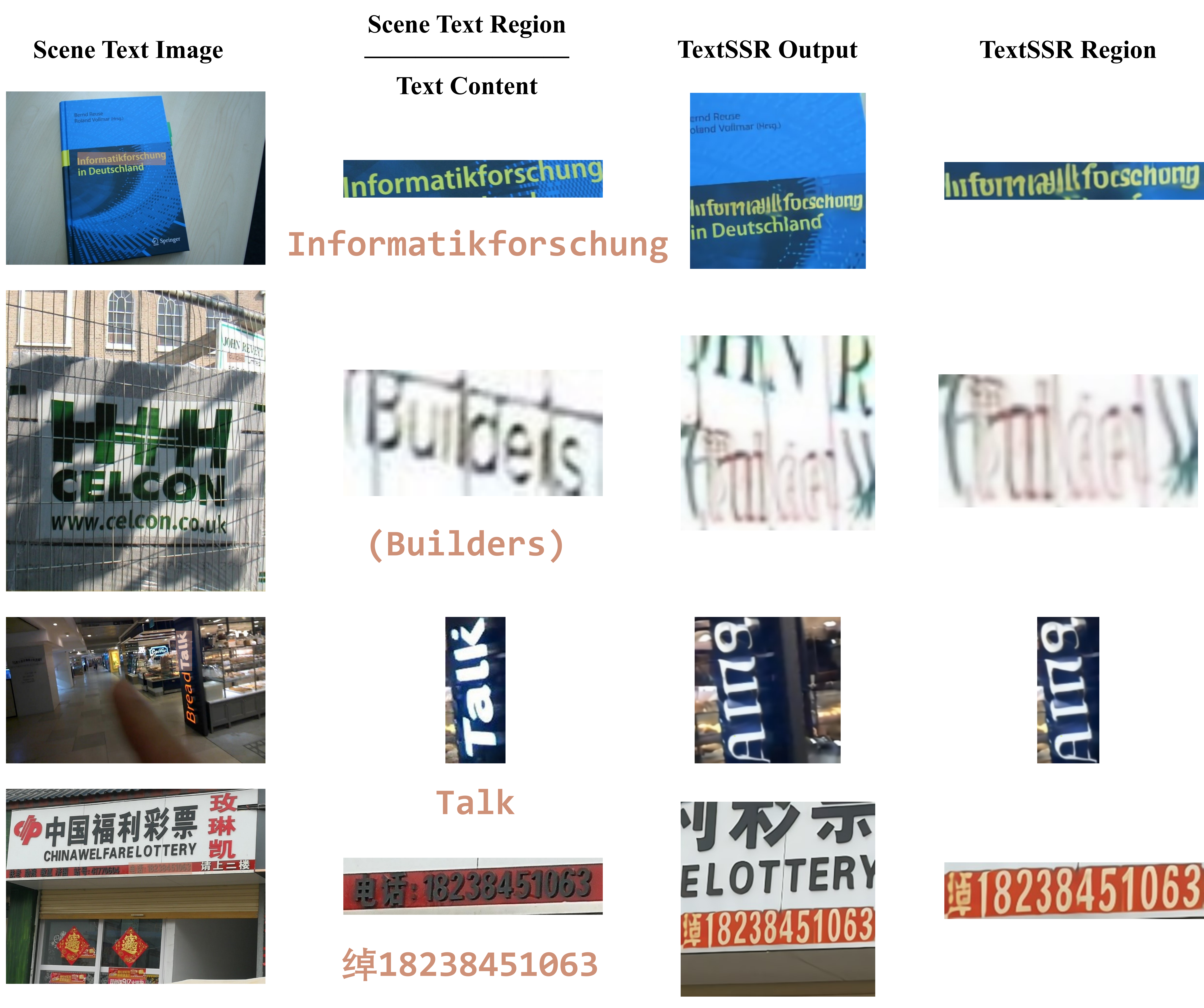}
    \caption{Failure Cases. We show several disappointing synthesis results produced by TextSSR.}
    \label{fig:failure}
\end{figure*}

It is important to note that our synthesis method is not flawless and has certain limitations. Fig.~\ref{fig:failure} presents several common failure cases, which can be attributed to the following reasons:
\begin{enumerate}
\item \textbf{Long text:} Excessively long text can confuse the model, resulting in disordered text images. This issue is exacerbated by the limited amount of training data for such cases.
\item \textbf{Blurred regions:} When the text region itself is excessively blurred, the model struggles to accurately reconstruct and synthesize the text.
\item \textbf{Multi-directional text:} The model, primarily trained on horizontally aligned text, faces challenges with multi-directional text, especially vertical text. Applying rotation-based post-processing, as used in STR methods, could be a potential solution.
\item \textbf{Incorrect text labels:} Errors in manual labeling can lead to mismatches between the rendered regions and their corresponding labels.
\item \textbf{Language Characteristics:} The performance on Chinese text is generally worse than on English, due to the higher number of characters and the complexity of Chinese characters.
\end{enumerate}

Despite the minority in quantity, handling challenging text instances are also important. We plan to tackle these instances as follows: splitting long text into shorter segments, simulating blur by adding noise and augmenting multi-directional text via rotation based on common instances, leveraging render-based data for pretraining on multilingual characters, etc.

\section{Discussion}

However, our study has limitations and avenues for further research, including the following: (1) The text location and the text content must be paired.  While we utilize the anagram-based method to mitigate this issue, we will design methods for reasonable, large-scale usable pairings for broader synthesis considerations.  (2) Currently, large-scale synthetic post-processing relies on an STR model;  we aim to integrate a self-checking mechanism into the entire framework to verify the correctness of the synthesized output.  This could further enhance learning and adjust the arrangement of text location until generating usable text correctly.   (3) Due to available large-scale scene text images already used for training, we plan to collect a larger dataset of untrained text images for the base images, creating a more extensive synthetic dataset to benefit the STR community. (4) Although the generation is related to the surrounding context, currently TextSSR does not fully address this issue due to lack of customized design. However, by substituting certain TextSSR component, e.g., the anagram expansion, or using LLM to recommend contextually appropriate content, TextSSR can largely alleviate it while the rest TextSSR components can still be reused. We will improve the semantic diversity and contextual realism from these aspects in future. (5) While our primary focus is on STR, our approach can also benefit other downstream tasks.  For example, by directly writing text onto the background or editing original text, our method can generate new data for text detection and document understanding tasks. We also agree that domain generalization is a valuable topic. We will investigate other downstream applications and discuss broader specialized domains in future.

\begin{figure*}[h!]
    \centering
    \includegraphics[width=0.9\linewidth]{imgs/more_viz.pdf}
    \caption{More visualization results for TextSSR.}
    \label{fig:more_viz}
\end{figure*}

\end{document}